\documentclass[a4paper, 10pt, conference]{ieeeconf}      

\IEEEoverridecommandlockouts                              

\usepackage{graphicx}
\usepackage{amsmath,amssymb,amsfonts}
\usepackage{algpseudocode,algorithm}
\usepackage{subcaption}
\usepackage{makecell}
\usepackage{cite}

\title{\LARGE \bf
Variational Adaptive Noise and Dropout \\towards Stable Recurrent Neural Networks
}

\author{Taisuke Kobayashi$^{1}$ and Shingo Murata$^{2}$
\thanks{*This work was supported by JST, CRONOS, Japan Grant Number JPMJCS24K6.}
\thanks{$^{1}$T. Kobayashi is with the National Institute of Informatics (NII) and with The Graduate University for Advanced Studies (SOKENDAI),
        2-1-2 Hitotsubashi, Chiyoda-ku, Tokyo, 101-8430, Japan
        {\tt\small kobayashi@nii.ac.jp}}%
\thanks{$^{2}$S. Murata is with the Department of Electronics and Electrical Engineering, Faculty of Science and Technology, Keio University, Japan
{\tt\small murata@elec.keio.ac.jp}}%
}

\begin{document}

\maketitle
\thispagestyle{empty}
\pagestyle{empty}

\begin{abstract}

This paper proposes a novel stable learning theory for recurrent neural networks (RNNs), so-called variational adaptive noise and dropout (VAND).
As stabilizing factors for RNNs, noise and dropout on the internal state of RNNs have been separately confirmed in previous studies.
We reinterpret the optimization problem of RNNs as variational inference, showing that noise and dropout can be derived simultaneously by transforming the explicit regularization term arising in the optimization problem into implicit regularization.
Their scale and ratio can also be adjusted appropriately to optimize the main objective of RNNs, respectively.
In an imitation learning scenario with a mobile manipulator, only VAND is able to imitate sequential and periodic behaviors as instructed.

\end{abstract}

\section{Introduction}

Deep learning for handling time-series data is important in robotics because robots operate continuously in the real world: for example, human motion prediction~\cite{aksan2021spatio}; robot motion generation~\cite{kim2021transformer}; and, in recent years, the understanding of natural language instructions~\cite{kawaharazuka2024real}.
The remarkable success in natural language processing has made Transformer-based models popular in such a time-series data processing recently~\cite{aksan2021spatio,kim2021transformer,kawaharazuka2024real}.
However, they are computationally expensive and are not suitable for modules that require real-time performance or where computational resources are limited.

Therefore, recurrent neural networks (RNNs) are still expected to be a useful deep learning technique for time-series data~\cite{hochreiter1997long,chung2014empirical}.
RNNs can be viewed as a kind of (learnable) dynamics in which the internal state output from a unit at one time is added to the unit's input at the next time, resulting in inference that reflects all past inputs, including the initial input (unless the effect on the internal state disappears).
The inputs must be given in chronological order, which makes parallelization difficult and makes it unsuitable for large-scale learning unlike Transformer, but it is computationally fast during inference and can be used in real-time even with limited resources on robots.
For example, continuous authentication based on the operator's operating habits was implemented using lightweight RNNs~\cite{kobayashi2022light}.
A world model incorporating RNNs has also been able to efficiently generate pseudo-experimental data, which can be used to learn the optimal policy for a quadruped robot~\cite{wu2023daydreamer}.

However, RNNs pose difficulties in learning.
RNNs typically optimize their dynamics by stochastic gradient descent (e.g.~\cite{ilboudo2023adaterm}), but RNNs can properly reflect past inputs in their outputs only if their gradients are back propagated in the past time direction, known as backpropagation through time (BPTT)~\cite{werbos1990backpropagation}.
Unfortunately, users of RNNs with BPTT would experience the unstable gradients as the computational graph for backpropagation grows in length over time~\cite{hochreiter2001gradient}.
For this reason, Truncated BPTT and its variants(e.g.~\cite{tallec2017unbiasing}), which cut off the computational graph at a certain time length, have been proposed, but they limits the past to be considered, which undermines the advantage of RNNs and also introduces a bias in learning.
In addition, as RNNs are a certain kind of dynamics, they must hold autonomous stability, namely, their internal state must not diverge when there is no input~\cite{erichson2021lipschitz}.
However, RNNs with gradient-based optimization cannot account for this constraint.

For these reasons, several studies have proposed the ways of stabilizing the learning of RNNs.
In particular, this paper focuses on the following two approaches.
Lim et al.~\cite{lim2021noisy} showed that learning can be stabilized by adding noise to the internal state.
Gal and Ghahramani~\cite{gal2016theoretically} stabilized RNNs by applying dropout, a technique to suppress overlearning of deep learning models, to the internal state, not to the input.
However, these noise and dropout were derived and examined separately, and their scale and ratio require fine-tuning.
This makes them impractical, and many users of RNNs prefer to decrease the learning rate for gradients than usual.
However, theoretically, this approach is not optimal, since it is known that generalization performance deteriorates as the learning rate is decreased~\cite{smith2018don}.

In this paper, therefore, we aim to develop a novel stable learning theory for RNNs.
Specifically, we reinterpret the optimization problem of RNNs as a stochastic model based on variational inference~\cite{hoffman2013stochastic}.
As a result, a noise is assigned to the internal state of RNNs.
We also derive a dropout by appropriately transforming the explicit regularization term introduced in the formulation into an implicit regularization~\cite{neyshabur2017implicit}.
The resulting noise scale and dropout ratio can be adaptively optimized to the optimization of the main objective using the reparameterization trick~\cite{kingma2014auto,liu2024bridging}.
The performance of the proposed method, so-called variational adaptive noise and dropout (VAND), is verified in an imitation scenario with a mobile manipulator.
The results show that VAND can stably learn to imitate the instructed sequential and periodic behaviors in two tasks with different time-series features, even under conditions where conventional learning methods fail.

\section{Preliminaries}

\subsection{Supervised learning for time-series data}

We first define the input and output data as $x \in \mathcal{X} \subset \mathbb{R}^{|\mathcal{X}|}$ and $y \in \mathcal{Y} \subset \mathbb{R}^{|\mathcal{Y}|}$, respectively.
A dataset is prepared (usually, in advance) as $N$ time-series trajectories $D=\{\tau_n\}_{n=1}^N$, where $\tau_n = \{(x_{n,t}, y_{n,t})\}_{t=1}^{T_n}$ with $T_n$ the sequence length.
A stochastic model $p$ with $\theta$ its parameters is learned in order to predict the output $y_{n,t}$ at time $t$ from the input $x_{n,t}$ and the previous input history $x_{n,<t}$ by minimizing the following negative log likelihood.
\begin{align}
    \theta^{\ast} = \arg \min_{\theta} \mathbb{E}_{\tau_n \in D} \left[- \sum_{t=1}^{T_n} \ln p(y_{n,t} \mid x_{n,t}, x_{n,<t}; \theta) \right]
    \label{eq:prob_general}
\end{align}
Note that $x_{n,<1}$ does not exist.
This minimization problem can be solved by stochastic gradient descent (in this paper, \cite{ilboudo2023adaterm}) on the replayed batch data with $B \leq N$ the batch size.

\subsection{Recurrent neural networks}

For the above problem, if the maximum sequence length $\max_n T_n$ is sufficiently short (or there are plenty of computing resources) and $x_{n,<t}$ for any $n=1,2,\ldots,N$ can be handled as is, models such as Transformer can be effective.
On the other hand, we need to consider situations such as online learning where the sequence length is not certain and ultimately infinite, or robot motion control where real-time performance is required with limited computational resources.
In that time, it is necessary to encode and approximate $x_{n,<t}$ into $H$-dimensional latent features (or internal state) $h_{n,t-1} \in \mathbb{R}^H$.
To obtain $x_{n,<t} \to h_{n,t-1}$, RNNs~\cite{hochreiter1997long,chung2014empirical} are popular models, which can be generally described as the following dynamics.
\begin{align}
    h_{n,t} = \mathrm{RNN}(x_{n,t}, h_{n,t-1}; \theta^\mathrm{RNN})
    \label{eq:def_rnn}
\end{align}
where $\theta^\mathrm{RNN}$ denotes the parameters of RNNs.
Note that it is common to set $h_{n,0} = 0$, corresponding to the point where $x_{n,<1}$ does not exist.
By optimizing $p(y_{n,t} \mid h_{n,t}; \theta)$, which replaces $(x_{n,t}, x_{n,<t})$ in eq.~\eqref{eq:prob_general} with the internal state $h_{n,t}$ of RNNs updated in this way, RNNs can handle arbitrary sequence lengths theoretically.

BPTT is widely used for training RNNs~\cite{werbos1990backpropagation}.
It considers the recurrent connections of RNNs to be multi-layer neural networks that are expanded in the time direction, and performs backpropagation on its computational graph.
This allows the current output to be optimized depending on past inputs.
However, with BPTT, the computational graph becomes large over time, and the gradient tends to disappear or diverge.
Therefore, to stabilize learning, the learning rate is often set smaller than in general deep learning models, but a small learning rate would cause poor generalization performance~\cite{smith2018don}.
Truncated BPTT, which limits the past time to be considered, is known to introduce bias~\cite{tallec2017unbiasing}.
Instead of these ad-hoc approaches, we need to theoretically improve the learning stability of RNNs.

\section{Proposal}

\subsection{Stability of RNNs}

\begin{figure}[tb]
    \centering
    \includegraphics[keepaspectratio=true,width=0.96\linewidth]{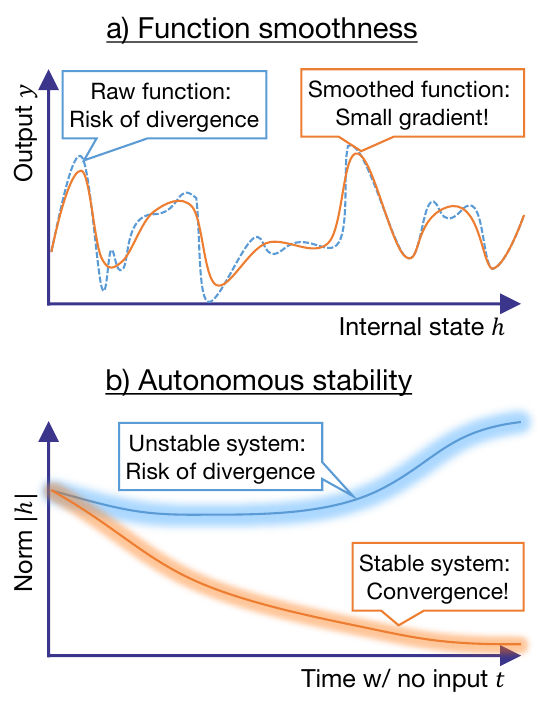}
    \caption{Stabilizing factors for RNNs}
    \label{fig:rnn_stability}
\end{figure}

Let's consider the factors for stabilizing the learning of RNNs, as illustrated in Fig.~\ref{fig:rnn_stability}.
First, the output of the function should vary smoothly with changes in the internal state of RNNs~\cite{lim2021noisy}.
The function smoothness makes the gradient less likely to diverge, and less likely to be extreme even for the chain rule of gradients in BPTT.
However, RNNs do not take it into account during design and optimization, leading to excessive gradients in the learning process.

On the other hand, RNNs are a kind of dynamics, so their learning stability in forward computation should be guaranteed by the autonomous stability with no input~\cite{erichson2021lipschitz}.
However, RNNs do not take it into account during design and optimization, leading to unstable systems in some times.
As a result, the numerically diversing  outputs cause excessive errors and make the learning of RNNs unstable.

Thus, to stabilize RNNs, the function smoothness and autonomous stability are required.
In this paper, we bring these two factors by rethinking the RNNs optimization problem, reinterpreting it as variational inference~\cite{hoffman2013stochastic} and converting its regularization term into implicit regularization~\cite{neyshabur2017implicit}.
Note that other stable methods for RNNs have been studied (e.g.~\cite{vorontsov2017orthogonality}), but since the proposed method is complementary to them, we omit a comparison with them in this paper.

\subsection{Formulation through variational lower bound}

From here, we omit the subscript $n$ for simplicity of description.
Instead of the minimization problem given in eq.~\eqref{eq:prob_general}, a model $p(y_t \mid x_t)$ is considered to stochastically predict the output $y_t$ from the input $x_t$ at time $t$.
In this model, $h_t$ is assumed to be hidden as a latent variable, yielding the following variational lower bound.
\begin{align}
    &\ln p(y_t \mid x_t)
    \nonumber \\
    =& \ln \mathbb{E}_{p(h_t \mid x_t)} \left[ p(y_t \mid h_t) \right]
    \nonumber \\
    =& \ln \mathbb{E}_{q(h_t \mid x_t, h_{t-1})} \left[ p(y_t \mid h_t) \frac{p(h_t \mid x_t)}{q(h_t \mid x_t, h_{t-1})} \right]
    \nonumber \\
    \geq& \mathbb{E}_{q(h_t \mid x_t, h_{t-1})}[\ln p(y_t \mid h_t)]
    \nonumber \\
    &- \mathrm{KL}(q(h_t \mid x_t, h_{t-1}) \| p(h_t \mid x_t))
\end{align}
where $p(h_t \mid x_t)$ is the prior distribution of $h_t$ conditioned on the latest input $x_t$ alone.
Although $q(h_t \mid x_t, h_{t-1})$ is a posterior distribution, the condition and target random variables indicate that it corresponds to RNNs.
In other words, the above maximization of the variational lower bound maximizes the log-likelihood of $p(y_t \mid h_t)$ after adding stochastic noise to the internal state, which is regularized to the prior distribution.
If the prior is designed to hold the autonomous stability, this regularization yields it as well.

\subsection{Convesion to implicit regularization}

To maximize the above variational lower bound for arbitrary time-series data and time, the first term is the same as eq.~\eqref{eq:prob_general}, except that the internal state of RNNs is with noise, while the regularization must be added explicitly as one of the loss functions.
Since this is not practical, we avoid this regularization by incorporating the implicit regularization corresponding to it into the forward computation of RNNs.

Specifically, we note that the second term can be transformed using a discriminator that identifies the generator of the random variable as follows~\cite{goodfellow2014generative}:
\begin{align}
    &\mathrm{KL} (q(h_t \mid x_t, h_{t-1}) \| p(h_t \mid x_t))
    \nonumber \\
    =& \mathbb{E}_{q(h_t \mid x_t, h_{t-1})} \left[ \ln \frac{q(h_t \mid x_t, h_{t-1})}{p(h_t \mid x_t)} \right]
    \nonumber \\
    =& \mathbb{E}_{q(h_t \mid x_t, h_{t-1})} \Biggl[ \ln \frac{q(h_t \mid x_t, h_{t-1})}{q(h_t \mid x_t, h_{t-1}) + p(h_t \mid x_t)}
    \nonumber\\
    &\quad\quad\quad\quad\quad\quad\times \frac{q(h_t \mid x_t, h_{t-1}) + p(h_t \mid x_t)}{p(h_t \mid x_t)} \Biggr]
    \nonumber \\
    =& \mathbb{E}_{q(h_t \mid x_t, h_{t-1})} \left[ \ln D(h_t)  - \ln (1 - D(h_t)) \right]
\end{align}
where
\begin{align}
    D(h_t) = \frac{q(h_t \mid x_t, h_{t-1})}{q(h_t \mid x_t, h_{t-1}) + p(h_t \mid x_t)}
\end{align}
Intuitively, $h_t$ generated from $q(h_t \mid x_t, h_{t-1})$ is misidentified as being generated from $p(h_t \mid x_t)$ when the two probability distributions match.
In other words, the strong regularization that minimizes the second term enough maximizes the probability of misidentification (at most 50~\%).

Considering the posterior distribution $\tilde{q}(h_t \mid x_t, h_{t-1})$ after regularization acts on the basis of this interpretation, it can simply be modeled as a mixture distribution of $q(h_t \mid x_t, h_{t-1})$ and $p(h_t \mid x_t)$.
\begin{align}
    &\tilde{q}(h_t \mid x_t, h_{t-1})
    \nonumber \\
    =& (1 - \beta) q(h_t \mid x_t, h_{t-1})
    + \beta p(h_t \mid x_t)
\end{align}
That is, the mixture coefficient is defined as $\beta \in [0, 1]^H$ (each element of the internal state is treated independently).
The larger $\beta$ is, the stronger the regularization is, and instead of adding the second regularization term explicitly to the loss functions to be minimized, $h_t$ can be generated from it to implicitly reflect the regularization.

To actually generate $h_t$ from $\tilde{q}(h_t \mid x_t, h_{t-1})$, it is necessary to concretize $p(h_t \mid x_t)$.
In this work, we assume $p(h_t \mid x_t) = q(h_t \mid x_t, 0)$, satisfying the following equation.
\begin{align}
    h_t &\sim \tilde{q}(h_t \mid x_t, h_{t-1}) = q(h_t \mid x_t, (1 - m) h_{t-1})
\end{align}
where $m \in \{0, 1\}^H$ is a mask, generated from an $H$-dimensional independent Bernoulli distribution $\mathcal{B}$ following $\beta$ as the dropout ratio.
This means that a dropout is performed to the past internal state when updating it in the computation of RNNs.
Note that the dropout is generally performed for all components of the internal state at once, but it is done for each component independently in this work.

\subsection{Implementation}

\begin{figure}[tb]
    \centering
    \includegraphics[keepaspectratio=true,width=0.96\linewidth]{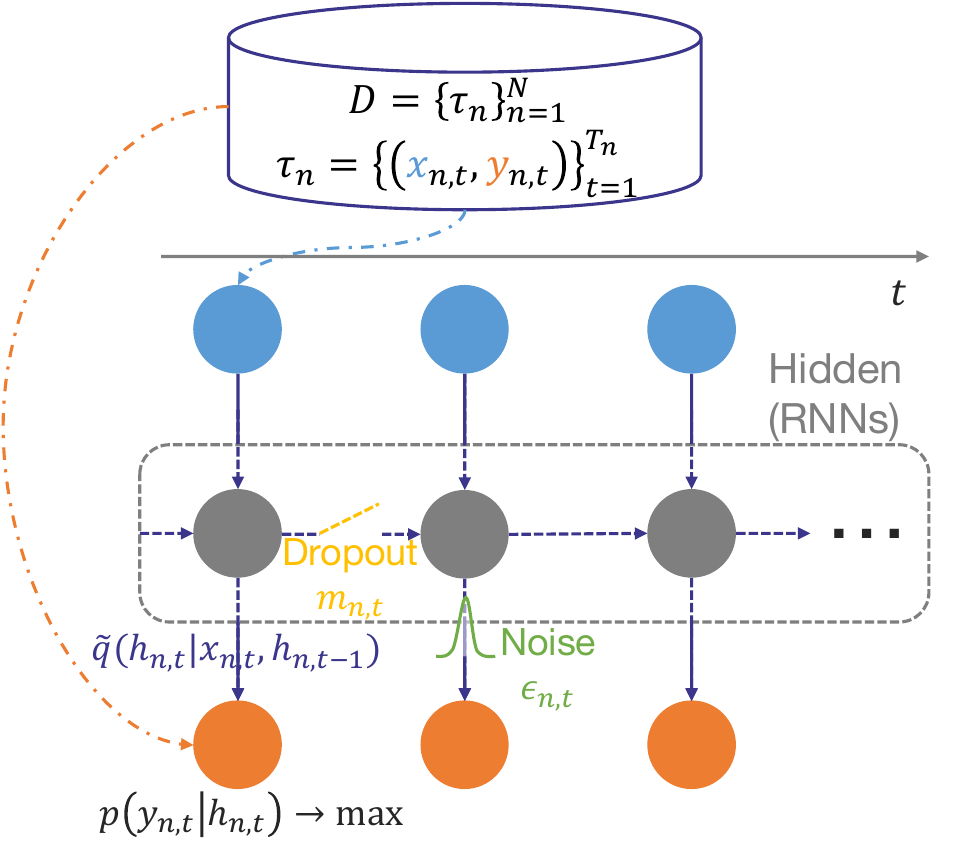}
    \caption{Proposed method: VAND}
    \label{fig:arch_vand}
\end{figure}

In summary, the proposed Variational Adaptive Noise and Dropout (VAND) method solves the following optimization problem for each time $t$ in the trajectory $\tau_n$ in the prepared dataset $D=\{\tau_n\}_{n=1}^N$ (see Fig.~\ref{fig:arch_vand}).
\begin{align}
    \theta^{\ast} =& \arg \min_{\theta} \mathbb{E}_{\tau_n \in D}\left[ - \sum_{t=1}^{T_n} \ln p(y_{n,t} \mid h_{n,t} + \epsilon_{n,t}; \theta) \right]
    \label{eq:prob_vand}
\end{align}
where
\begin{align*}
    h_{n,t} &= \mathrm{RNN}(x_{n,t}, (1 - m_{n,t}) h_{n,t-1}; \theta^\mathrm{RNN})
    \\
    \epsilon_{n,t} &\sim \mathcal{N}(0, \sigma)
    ,\
    m_{n,t} \sim \mathcal{B}(\beta)
\end{align*}
where $\sigma \in \mathbb{R}_+^H$ is the noise scale for Gaussian $\mathcal{N}$ (representing the posterior distribution).
$\sigma$ and $\beta$ can be optimized simultaneously when solving the above optimization problem by making full use of the reparameterization trick~\cite{kingma2014auto}.
That is, the proposed VAND imposes the adaptive noise and dropout on RNNs, which should maximize the prediction ability.
At the inference mode, the above sampled variables are replaced by the means of their probability distributions, i.e. $\epsilon_{n,t} = 0$, $m_{n,t} = \beta$.

As a remark, because both $\sigma$ and $\beta$ are domain bounded, their updates are suppressed near their boundaries.
To alleviate this problem, the following gradient operation corresponding to the mirror descent algorithm and straight-through backpropagation~\cite{liu2024bridging} is performed.
\begin{align}
    \sigma &= \mathrm{softplus}(\hat{\sigma}^\mathrm{real}) + \sigma^\mathrm{real} - \hat{\sigma}^\mathrm{real}
    \\
    \beta &= \mathrm{sigmoid}(\hat{\beta}^\mathrm{real}) + \beta^\mathrm{real} - \hat{\beta}^\mathrm{real}
\end{align}
where $\sigma^\mathrm{real} \in \mathbb{R}^H$ and $\beta^\mathrm{real} \in \mathbb{R}^H$ denote the parameters before transformation by $\mathrm{softplus}$ for positiveness and $\mathrm{sigmoid}$ for boundedness, respectively.
$\hat{\cdot}$ is the stop-gradient operation to cut its computational graph while maintaining its numerical value.

Adding noise to the process during training makes the loss function to be optimized smooth, making it harder for the gradients to diverge during backpropagation.
The dropout also has the effect of decaying the internal state by a percentage $\beta$, indicating that RNNs are more likely to achieve autonomous stability.
In the previous studies~\cite{lim2021noisy,gal2016theoretically}, RNNs are stabilized by introducing pre-designed noise scale and dropout ratio separately, but the proposed method introduces both simultaneously and in a form that can be adjusted automatically, expecting greater stabilization than the previous studies (without fine-tuning).
Note that although they require additional computational cost, the computational complexity is the same as for the conventional RNNs, both during inference and training.

\section{Experiments}

\subsection{Tasks}

\begin{figure}[tb]
    \centering
    \includegraphics[keepaspectratio=true,width=0.84\linewidth]{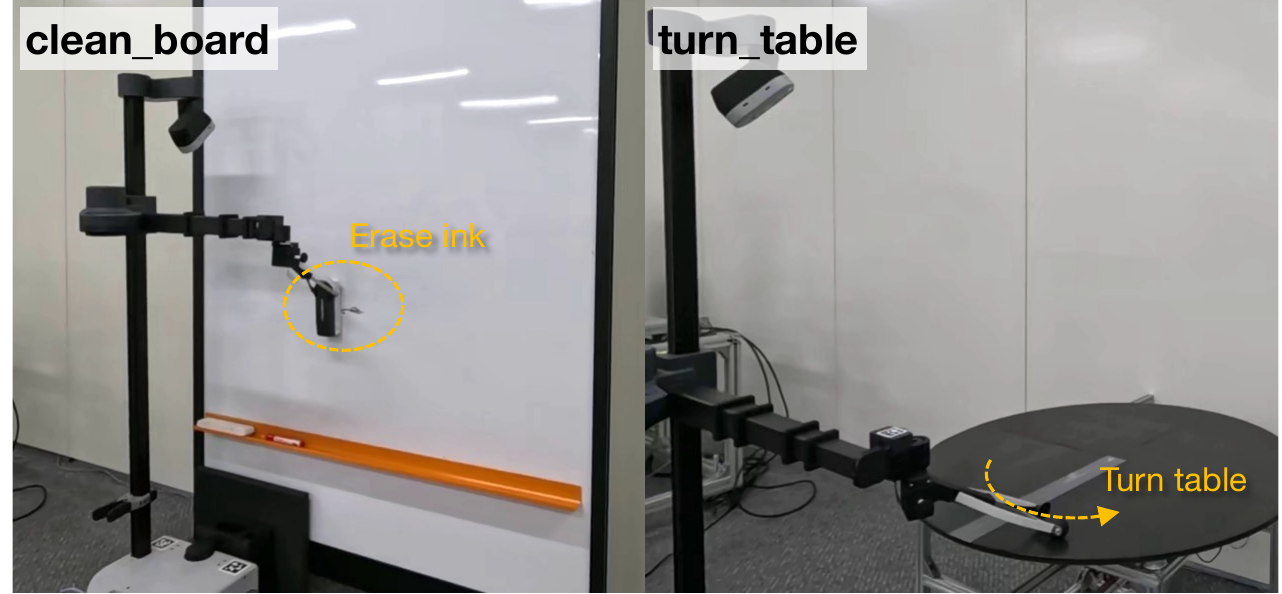}
    \caption{Experimental setup}
    \label{fig:exp_setup}
\end{figure}

We evaluate the effectiveness of the proposed VAND through two imitation learning tasks.
An operator controls Hello Robot's mobile manipulator Stretch 1 with a joystick to conduct the following tasks (see Fig.~\ref{fig:exp_setup}):
in the \textit{clean\_board} task, the robot grasps an eraser to erase ink on a whiteboard;
and in the \textit{turn\_table} task, the robot touches a rotatable table and repeatedly turn it counterclockwise.
Since the former must learn a long-term process, the addition of noise is desired for getting the function smoothness with robustness to internal state fluctuations.
On the other hand, the latter must learn a stable limit cycle, so the addition of dropout is desired for getting the autonomous stability.

In imitation learning~\cite{bain1995framework}, the robot predicts the operator's commands (or actions) from the robot's observations as inputs.
The observations are 34 dimensions: position, velocity, and effort of the robot arm, wrist, and gripper; position and velocity of the base; acceleration of the IMU mounted on the wrist; acceleration and gyro of the base IMU; and the command values executed before.
Note that since the base is not moved in the \textit{turn\_table}, we simplified it to 19 dimensions by excluding the 15 dimensions related to the base.
The actions in \textit{clean\_board} are 4 dimensions: 2D movement of the arm; the opening and closing of the gripper; and the translation of the base.
In \textit{turn\_table}, the actions are 3 dimensions: 2D movement of the arm; and the wrist rotation.

The training data for both tasks consisted of 10 trajectories of 600 steps (i.e. approximately 30 seconds).
However, as part of the data augmentation, the initial time of the trajectory was randomly selected from 1 to 10 steps, and the end time was reduced accordingly to align the trajectory length.
The optimization of eq.~\eqref{eq:prob_vand} was performed using stochastic gradient descent with a batch size of 50, and the performance of the model was evaluated after a total of 1000 epochs.
The test data used for this performance evaluation consists of 4 similar trajectories for \textit{clean\_board} and two trajectories with 1200 steps (i.e. about 60 seconds) for \textit{turn\_table}.
The MSE scores for these are compared (i.e. smaller is better) for the learning conditions described next.

\subsection{Conditions}

The above tasks are trained on a common model consisting of two layers of LSTM~\cite{hochreiter1997long} with 100 units for each\footnote{The trend does not change significantly for other models of RNNs (e.g.~\cite{chung2014empirical}).}.
This model takes the observations and outputs the action's normal distribution (i.e. its mean and variance).
As a noise-robust optimizer, AdaTerm~\cite{ilboudo2023adaterm} is employed with its default setting.
Its learning rate is $10^{-3}$, which is large for RNNs, so instability in learning is expected.

There are six comparisons in total, depending on the presence or absence of noise and dropout and their optimization.
\begin{itemize}
    \item \textit{Vanilla}: the case without no noise and dropout as the standard RNNs
    \item \textit{c/v-Noise}: the case only with noise, the scale of which is constant (`c`) or variable (`v`)
    \item \textit{c/v-Dropout}: the case only with dropout, the ratio of which is constant (`c`) or variable (`v`)
    \item \textit{VAND}: the proposed method with both noise and dropout, which are optimized
\end{itemize}
Note that when the noise scale and dropout ratio are constants, they are given at their lowest value in implementation, $10^{-2}$.
When they are optimized as variables, their initial values are given corresponding to $\sigma^\mathrm{real}=0$ and $\beta^\mathrm{real}=0$, respectively.
Anyway, these six comparisons are trained 20 times with different random seeds and their performances are compared statistically.

\subsection{Results}

\begin{figure}[tb]
    \centering
    \includegraphics[keepaspectratio=true,width=0.92\linewidth]{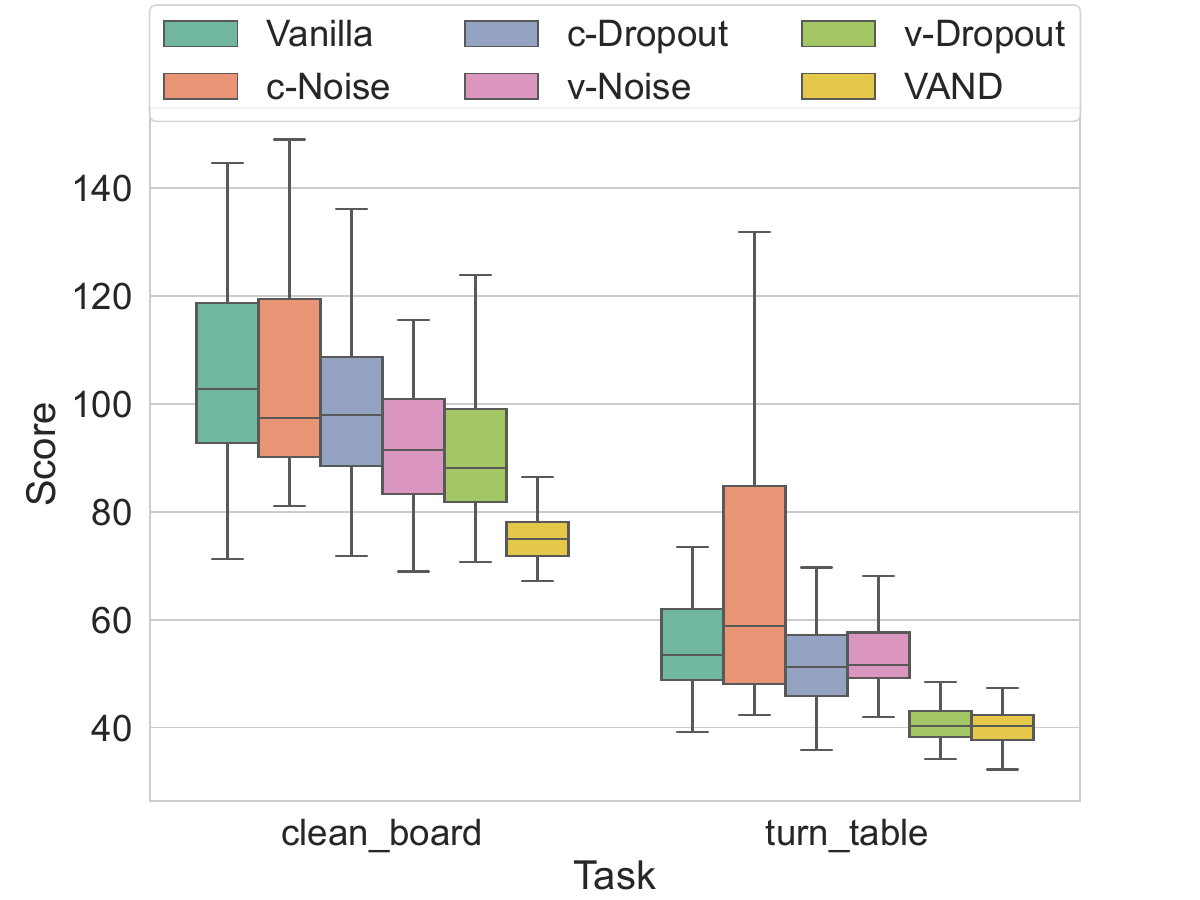}
    \caption{Test MSE of six comparisons on two tasks}
    \label{fig:test_mse}
\end{figure}

The learning results are depicted in Fig.~\ref{fig:test_mse}.
As can be seen at a glance, the proposed VAND stably achieved the smallest MSE for both tasks.
The conventional RNNs training, Vanilla and the constant versions of Noise and Dropout, could not learn to imitate properly (without fine-tuning).
On the other hand, the adaptive noise and/or dropout improved performance.
In particular, the optimized noise greatly reduced the worst-case scenario for the \textit{clean\_board}, and the optimized dropout for the \textit{turn\_table} yielded a satisfactory imitation performance.
These results are in line with initial expectations, as the function smoothness prevents the occurrence of extreme gradients and the autonomous stability leads to convergence to stable limit cycles.

\begin{figure}[tb]
    \begin{subfigure}[b]{0.84\linewidth}
        \centering
        \includegraphics[keepaspectratio=true,width=\linewidth]{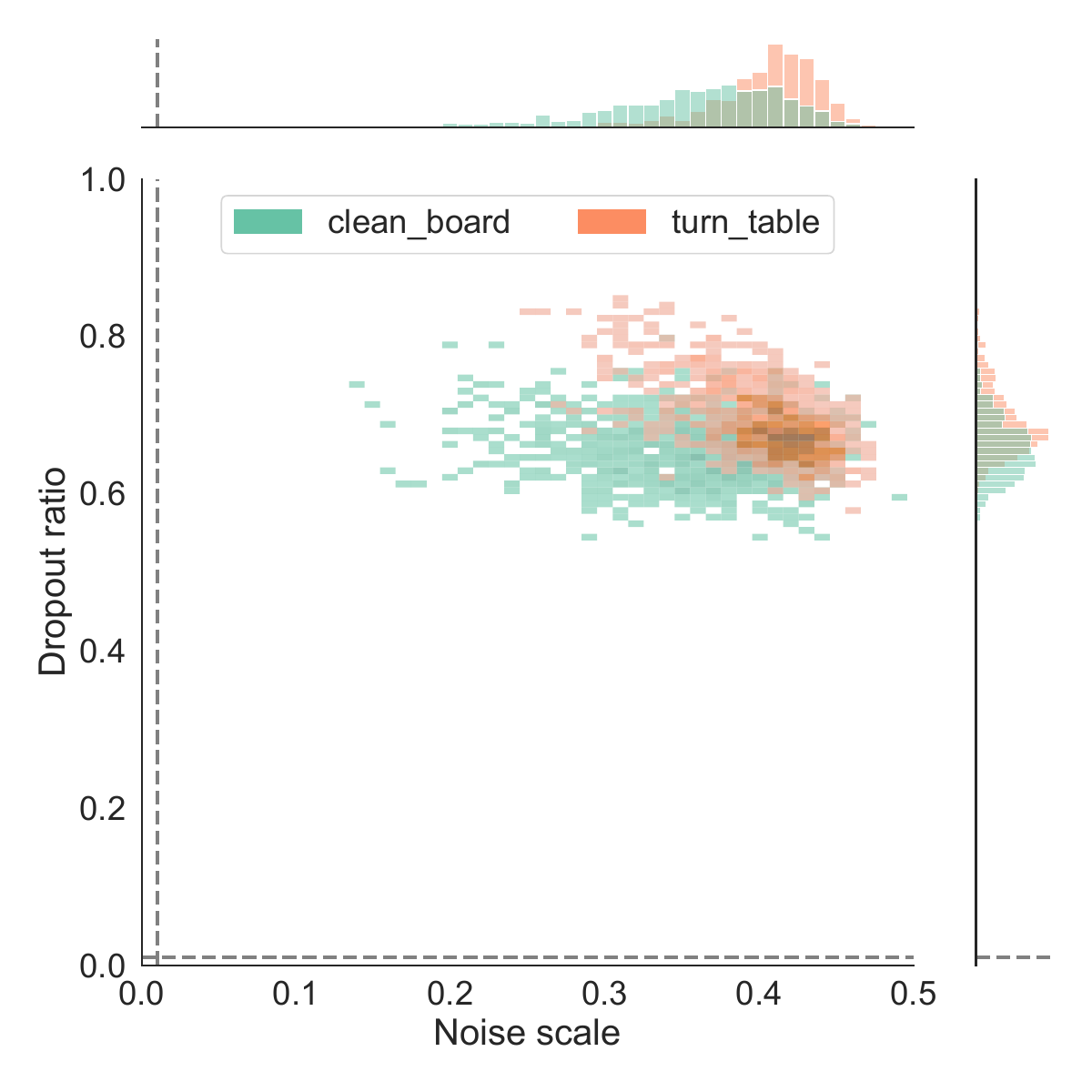}
        \subcaption{First layer (input side)}
        \label{fig:analysis_1}
    \end{subfigure}
    \begin{subfigure}[b]{0.84\linewidth}
        \centering
        \includegraphics[keepaspectratio=true,width=\linewidth]{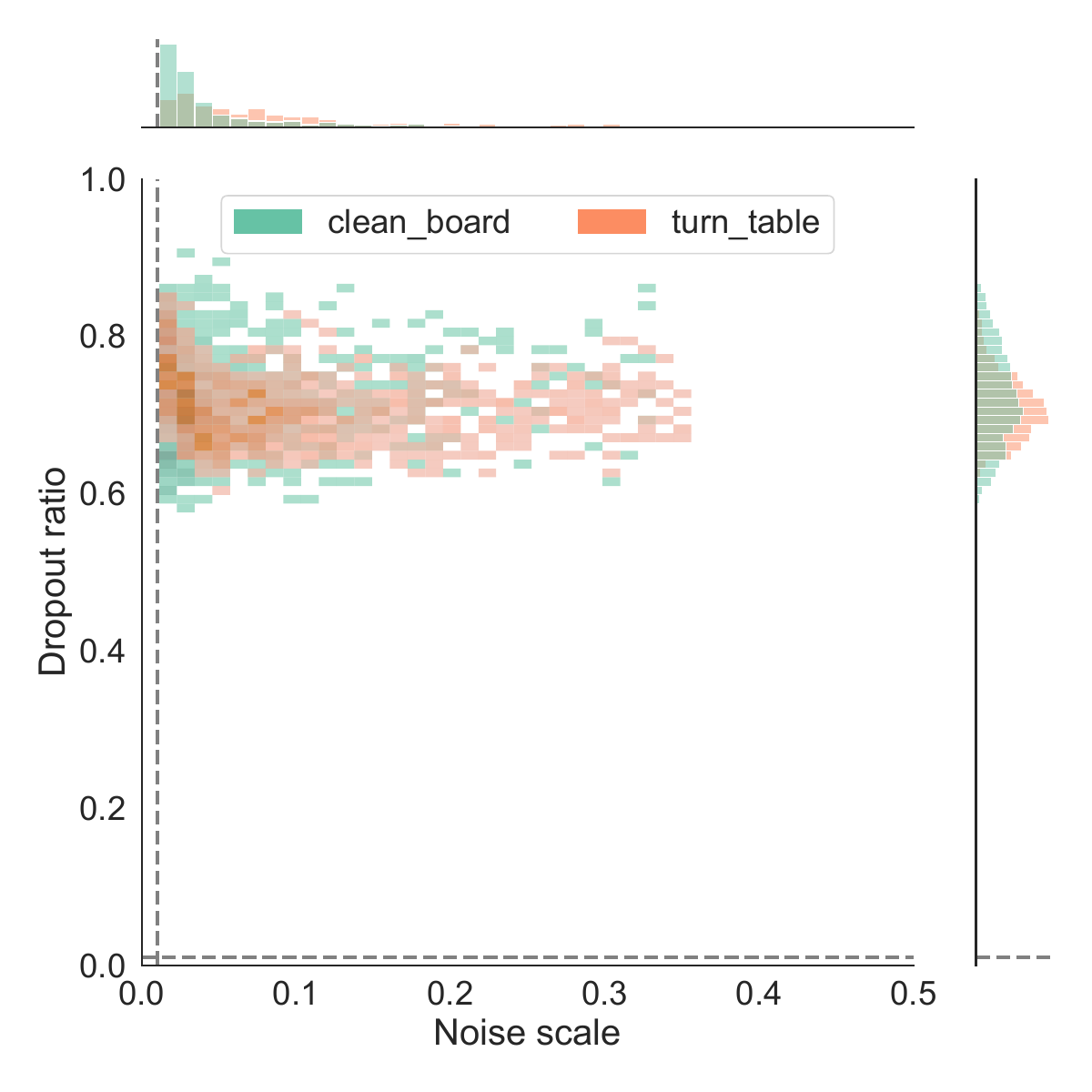}
        \subcaption{Second layer (output side)}
        \label{fig:analysis_2}
    \end{subfigure}
    \caption{Joint plots of the noise scale and dropout ratio in the models after training}
    \label{fig:analysis}
\end{figure}

Although it was found that the adaptive noise and dropout contribute to imitation performance, it is necessary to confirm that they are appropriately tuned according to tasks and implementations.
Therefore, the joint plots of the noise scale and dropout ratio of the models after training is depicted in Fig.~\ref{fig:analysis}.
The obtained plots have different shapes depending on the tasks and layers.
In particular, the stabilizing factors were stronger in the first layer than in the second layer, suggesting that the instability due to multi-layer structure was more dominant than the performance improvement in the two tasks.
In addition, the noise scale and dropout ratio tended to be smaller for \textit{clean\_board} than for \textit{turn\_table}, indicating that they were tuned to achieve the long-term prediction performance required by \textit{clean\_board} and to achieve the periodic stability required by \textit{turn\_table}.

\subsection{Demonstrations}

\begin{figure}[tb]
    \begin{subfigure}[b]{0.84\linewidth}
        \centering
        \includegraphics[keepaspectratio=true,width=\linewidth]{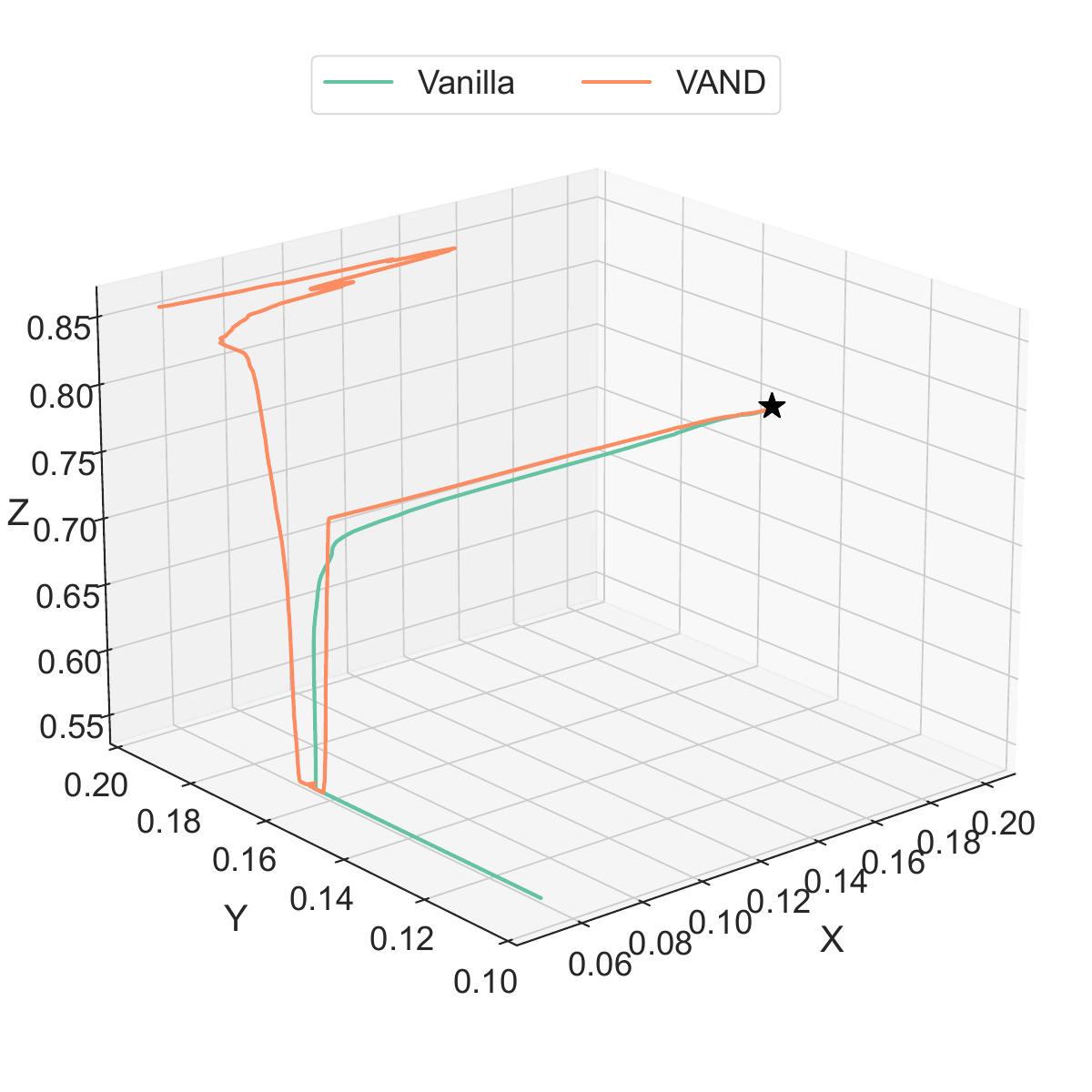}
        \subcaption{\textit{clean\_board}}
        \label{fig:traj_clean_board}
    \end{subfigure}
    \begin{subfigure}[b]{0.84\linewidth}
        \centering
        \includegraphics[keepaspectratio=true,width=\linewidth]{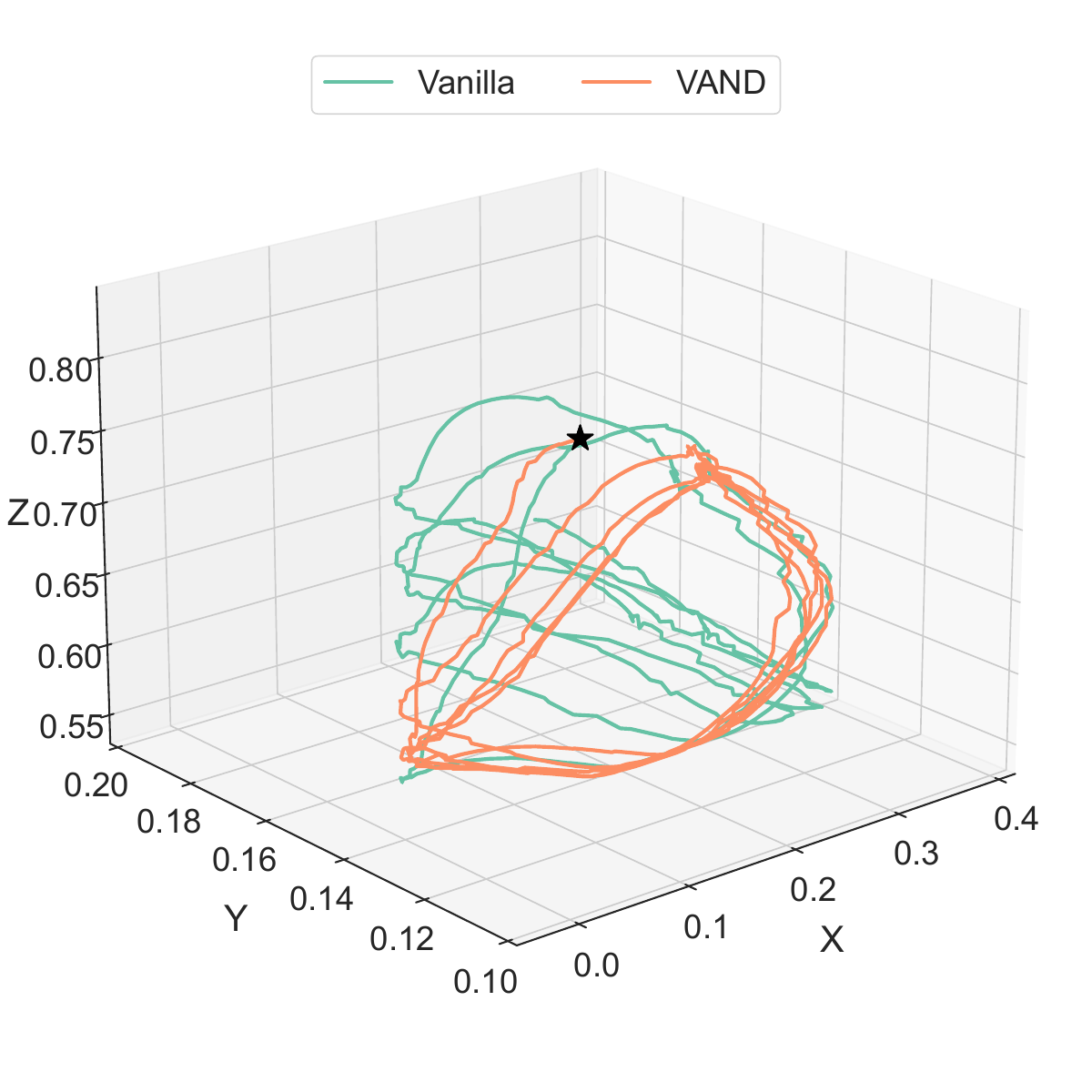}
        \subcaption{\textit{turn\_table}}
        \label{fig:traj_turn_table}
    \end{subfigure}
    \caption{Trajectories starting from a star by the conventional and proposed methods on two tasks}
    \label{fig:traj}
\end{figure}

The imitation performance of the learned model is demonstrated.
Here, we test the conventional method, Vanilla, and the proposed method, VAND.
Among 20 trials of them, the models with the smallest test MSE are employed.
The trajectories of the end effector are illustrated in Fig.~\ref{fig:traj}, and the corresponding motions are shown in the attached video.

First, Vanilla clearly failed to learn in \textit{clean\_board}, and the robot moved backward at the timing of grasping the eraser.
This was likely due to the excessive influence of the positioning behavior included in the dataset, which caused the robot to move backward a bit and continue due to the distribution shift.
On the other hand, VAND was able to execute the sequential process according to the instructed procedure.
The same was true for the \textit{turn\_table}, where Vanilla could hardly turn the table, in addition to the trajectory of the periodic motion being significantly off each time.
In contrast, VAND quickly converged to a stable limit cycle and was able to stably turn the table as instructed.

\section{Conclusion}

In this paper, we proposed a novel method, VAND, to stabilize the learning of RNNs on time-series data.
It reinterprets the optimization problem of RNNs in the framework of variational inference with the implicit regularization, resulting in the adaptive noise and dropout to the internal state of RNNs.
In an imitation learning scenario with a mobile manipulator, we performed two tasks with different time-series features.
As a result, the proposed VAND successfully imitated the instructed sequential and periodic behaviors despite conditions where conventional methods failed.

In VAND implemented in this paper, the noise scale and dropout ratio are given to be learnable variables, but there is room to consider designing them as functions that depend on the input and past internal state, such as the gate mechanism in LSTM.
It is also possible that it would be useful to structure them appropriately, rather than optimizing them only for the main objective that RNNs want to solve.

%
%
\bibliographystyle{IEEEtran}
\bibliography{biblio}

\end{document}